\documentclass[journal]{IEEEtran}
\usepackage{blindtext}
\usepackage{graphicx}
\usepackage{color}
\usepackage{gensymb}
\usepackage{amsmath}
\usepackage{amssymb}

\ifCLASSINFOpdf
\else
\fi
\begin{document}
%\title{A Real-time  Query Search Algorithm for Video}

%\title{A Body Feature-based Online Query Scheme for Instantly Interactive Edge Surveillance Systems}

\title{I-ViSE: Interactive Video Surveillance as an Edge Service using Unsupervised Feature Queries}

\author{
\IEEEauthorblockN{Seyed Yahya Nikouei${^a}$, Yu Chen${^a}$, Alexander Aved$^{b}$, Erik Blasch$^{b}$}

\IEEEauthorblockA{${^a}$Dept. of Electrical \& Computer Engineering, Binghamton University, SUNY,  Binghamton, NY 13902, USA \\ $^{b}$The U.S. Air Force Research Laboratory, Rome, NY 13441, USA\\
\{snikoue1, ychen\}@binghamton.edu, \{alexander.aved, erik.blasch\}@us.af.mil}}

\maketitle

\begin{abstract}

Situation AWareness (SAW) is essential for many mission critical applications. However, SAW is very challenging when trying to immediately identify objects of interest or zoom in on suspicious activities from thousands of video frames. This work aims at developing a queryable system to instantly select interesting content. While face recognition technology is mature, in many scenarios like public safety monitoring, the features of objects of interest may be much more complicated than face features. In addition, human operators may not be always able to provide a descriptive, simple, and accurate query. Actually, it is more often that there are only rough, general descriptions of certain suspicious objects or accidents. This paper proposes an Interactive Video Surveillance as an Edge service (I-ViSE) based on unsupervised feature queries. Adopting unsupervised methods that do not reveal any private information, the I-ViSE scheme utilizes general features of a human body and color of clothes. An I-ViSE prototype is built following the edge-fog computing paradigm and the experimental results verified the I-ViSE scheme meets the design goal of scene recognition in less than two seconds.

\end{abstract}


\IEEEpeerreviewmaketitle

\section{Introduction}
\label{sec:intro}

Smart Cities pervasively deploy video cameras for information collection \cite{xu2018real} and Situation Awareness (SAW). While cameras enable 24-7 continuous collection of city footprints, the huge amount of video data brings new challenges, among which the top two are the scalability \cite{nikouei2018real} and the privacy \cite{fitwi2019lightweight}. As streaming video increases, it becomes infeasible to have human operators sitting in front of hundreds of screens to catch suspicious activities or identify objects of interests in real-time. Actually, with millions of surveillance cameras deployed, \emph{video search} is more vital than ever. For example, it is very time consuming for the operator to find a specific scene where a certain action took place among hundreds of hours of video streams. As from the experimental collection, when a security officer is looking for a suspicious person on the run, the cameras are not adequately responsive. In collection scenarios, a method that allows real-time video querying and facilitates thousands of frames and performs instant object identification is desperately needed, which is able to look through thousands of frames and identify the object of interest instantly. Meanwhile, many people are very much concerned, some are even paranoid about the invasion of their privacy by the cameras from streets, stores, and in the community \cite{cavallaro2007privacy}.

%This work aims at an interesting but challenging question that is more meaningful to highly interactive operations: \emph{can we make the security surveillance systems queryable such that it is able to selectively play the contents of interest instantly for mission-critical tasks?} 

This work aims at enhancing security surveillance through efficient design of queryable operations. The query responses selectively highlights meaningful content and instantly provides interactive knowledge of mission-critical tasks. The paper tries to answer an interesting but challenging question that is very meaningful to highly interactive operations: \emph{can we make surveillance systems queryable and privacy-preserving?}

An ideal security surveillance algorithm is expected to fulfill the following functions without violating people's privacy: (1) identify the object of interest, (2) match the video frames with the description query, and (3) report the camera ID or geo-location. Although face recognition based approaches are very mature today, it brings up deep concerns on privacy violation. In many practical application scenarios like public safety monitoring, features of objects of interest may be much more complicated than facial feature recognition. In addition, the operators may not be always able to provide simple, concise, and accurate queries. Actually, it is more often that operators merely provide rough, general, and uncertain descriptions of certain suspicious objects or accidents. %where the matching frames are shot. Additionally, in many practical application scenarios, the features of objects of interest may be much more complicated than face features and the operators may not be always able to provide a very simple, accurate query. It is more often that there are merely some rough, general descriptions of certain objects or accidents. 

Because of the tight constraints on time delays and communication network bandwidth, it is not practical to outsource the huge amount of raw video streams to a cloud center to instantly process the queries. Instead, edge computing is a promising solution. Edge computing allows computational tasks conducted by smart Internet of Things (IoT) devices on-site or near-site, which enables instant information procession and decision-making \cite{nikouei2019kerman}. In addition, the novel microservices architecture, a variant of the service-oriented architecture (SOA) structural style, supports development of lightweight applications for the edge environment as a collection of loosely coupled, fine-grained applications \cite{yu2018survey}.

%Because of the real-time requirement and the huge amount of cameras, it is not practical to outsource all the raw video streams to a cloud center for instant query processing. It also incurs unnecessary workload to the communication networks. Edge computing is promising. It allows tasks conducted by smart Internet of Things (IoT) devices on-site or near-site, which enables instant information procession and decision-making \cite{nikouei2019toward}. In addition, the novel microservices architecture, a variant of the service-oriented architecture (SOA) structural style, supports development of lightweight applications fitting in the edge environment \cite{yu2018survey}. Taking advantages of the microservices architecture, the functionality can be structured as a collection of loosely coupled, fine-grained services. 

This paper proposes an Interactive Video Surveillance as an Edge service (I-ViSE) based on unsupervised queries, which allows the operator to search by keywords and feature descriptions. The I-ViSE system matches query searches with captured video frames where the objects of interest appear. The I-ViSE search platform gives the option to utilize a set of microservices to look for features in a mathematical model such as objects, people, color, and behaviors. Adopting unsupervised classification methods, the I-ViSE scheme works with the general features such as a human body and color of clothes, while not violating the privacy of residents being monitored. The I-ViSE prototype is built following the edge-fog computing paradigm and the experimental results verify the I-ViSE scheme meets the real-time requirements. In summary, the contributions of I-ViSE can be itemized as follows:

\begin{itemize}
    \item[1)] To the best of our knowledge, this is the first real-time video querying algorithm that searches objects of interest using an unsupervised method, which does not violate people's privacy;
    
    \item[2)] A microservices architecture design within the edge hierarchy platform is introduced, which makes the query management algorithm lightweight and robust;
    
     \item[3)] An unsupervised training method is proposed that accurately matches the query to the pixel blob; and
    
    \item[4)] A prototype is implemented using Raspberry Pi verifying the effectiveness of the decentralized query method in terms of delay, resource consumption, and the detection accuracy.
\end{itemize}

The rest of this paper is structured as follows. Section \ref{sec:back_know} provides a brief overview of the related work and the background knowledge. Section \ref{sec:platform} introduces the I-ViSE system design. Section \ref{sec:frame} presents the query processing scheme and delivery of results. Section \ref{sec:search} discusses the online query working flow details. The experimental results are analyzed in Section \ref{sec:experimental} and finally conclusions are presented in Section \ref{sec:conclusion}.
%%%%%%%%%%%%%%%%%%%%%%%%%%%%%%%%%%%%%%%%%%%%%%%%%%%%%%%%%%%%%%%%%%%%%%%%%%%%%%%%%%%%%%%%%%%%%%%%%%%%%%%%%%%%%%%%%%%%%%%%%%%%%%%%%%%%%%%%%%%%%%%%%%%%%%%%
\section{Background Knowledge}
\label{sec:back_know}

\subsection{Microservices}

The traditional service-oriented architecture (SOA) is monolithic, constituting different software features in a single interconnected database and interdependent applications. While the tightly coupled dependence among functions and components enables a single package, such a monolithic architecture lacks the flexibility to support continuous development and streaming data delivery, which is critical in today’s quickly changing and highly heterogeneous environment. 

Microservices architecture has been adopted to revitalize the monolithic architecture based applications, including the modern commercial web application. The flexibility of microservices enables continuous, efficient, and independent deployment of application function units. Significant features of microservices include fine granularity, which means each of the microservices can be developed in different frameworks like programming languages or resources, and loose coupling where the components are independent of function deployment and development  \cite{yu2018survey}.

A microservices architecture has been investigated in smart solutions to enhance the scalability and security of applications. It was used to implement an intelligent transportation system that incorporates and combines IoT to help planning for rapid bus systems \cite{herrera2018smart}. In another application, the microservices architecture was used to develop a smart city IoT platform where each microservice is regarded as an engineering department. The independent behavior of each microservice allows flexibility of selecting the development platform, and the communication protocols are simplified without requiring a middleware \cite{krylovskiy2015designing}.

\subsection{Human Keypoint Detection}

Human pose estimation, which refers to the keypoint detection on the body of the subject, is a long standing research area. Using manually selected features \cite{yang2011articulated}, \cite{wang2013beyond} is insufficient in locating the body parts effectively. More recent convolutional approaches, in which the machine extracts the features, drastically improved performance. There are two main methods of human recognition including a single person pipeline or the multi-person pipeline \cite{dang2019deep}. The single person category is further divided to heat-map generation where each pixel shows the probability of a likely keypoint. The heat map examples derive from Generative Adversarial Networks (GAN) \cite{goodfellow2014generative}, ``Stacked Hourgalss'' model \cite{newell2016stacked}, or Convolutional Pose Machines (CPM) \cite{wei2016convolutional}. Another approach is regression on the feature map to the keypoint locations \cite{carreira2016human}, \cite{sun2017compositional}. 

The multi-person detection pipeline can be divided into Top-down approaches and Bottom-up approaches. Top-down methods detect each person and then recognize each person’s keypoints \cite{chen2018cascaded}, \cite{fang2017rmpe}, \cite{iqbal2016multi}. Bottom-up methods have reversed order of steps: the first step is to locate all the keypoints in an image and then to group these keypoints according to the person they belong to \cite{cao2017realtime} and \cite{zhu2017multi}. Recently, researchers also tried to find the whole body estimation using only a single network \cite{hidalgo2019single}, which improves the performance drastically compared to the %\textcolor{red}{
well-known OpenPose \cite{cao2018openpose}%}
. The model uses VGG19 architecture for convolution filter layers.

\subsection{Video Query}

There are many efforts that use the Deep Neural Networks (DNNs) to make sense of video and present the labels for query purposes. Labels can be searched for in semi-real-time or they may be indexed for future references. A model discussed in \cite{ananthanarayanan2017real} can be used for video analysis to track objects using an edge system for a better understanding of urban intersections on how the cars and pedestrians behave. Similar approaches are suggested using DNN to summarize the video \cite{xu2016discovery} such as the street extraction, car detection, and path determination. Another example uses Natural Language Processing (NLP) techniques adopted with a CNN to give sentences of actions in video segments  \cite{wang2019asymmetric}. Researchers also introduced a method to compare video segments, available in a public data set MVS1K, where images searched by a query on the web are used as preferences for query intent \cite{ji2019query}. 

More recently, the research community has turned its focus to deploy detection models to better search in the videos using query engines. After parsing an image, the engine looks at tables that are filled with the detection results from the video processing algorithms \cite{kang2019challenges}. In 2018, a position paper introduced a distributed network that is capable of accepting queries in both real-time and an indexed version for video analysis \cite{nagothu2018microservice}. Following these works, we investigate a distributed version of a query language for video search and index the features for faster off-line analytical searches. 

%%%%%%%%%%%%%%%%%%%%%%%%%%%%%%%%%%%%%%%%%%%%%%%%%%%%%%%%%%%%%%%%%%%%%%%%%%%%%%%%%%%%%%%%%%%%%%%%%%%%%%%%%%%%%%%%%%%%%%%%%%%%%%%%%%%%%%%%%%%%%%%%%%%%%%%%
\section{I-ViSE Scheme Overview}
\label{sec:platform}

%Video query is a broad topic which is relatively untouched by the smart surveillance community. The concept of video analysis is limited to object detection and classification of pictures, which actually can be considered as the first step toward understanding of the video data. Attempts to query visual data uses the deep learning models to classify frames with the specific objects that are in each frame while giving the bounding box around each one. The I-ViSE enables the security personnel conduct real-time search in a large scale smart surveillance system based on some high-level, not-so-accurate descriptions on the object of interest. For instance, the phrases like ``red hoodie, blue jean'' are applicable as the keys and the I-ViSE system returns the matches with geolocation associated with the cameras. %Also, the platform is capable of indexing all of the frames that are processed in the fog node connected to the edge. The connections are based on HTML. 

I-ViSE uses video query for smart urban surveillance. The first step toward understanding of the video data begins with object detection and classification of pictures.  Visual data querying uses the deep learning models to classify specific objects in frames with bounding boxes. The I-ViSE enables the security officers to conduct real-time search in a large scale smart surveillance system based on high-level, not-so-accurate descriptions on the object of interest. For instance, the phrases like “red hat, blue jeans” are normally applicable as the keys and the I-ViSE system returns the matches with geolocation associated with the cameras.

\begin{figure}[t]
    \centering
        \includegraphics[width=0.42\textwidth]{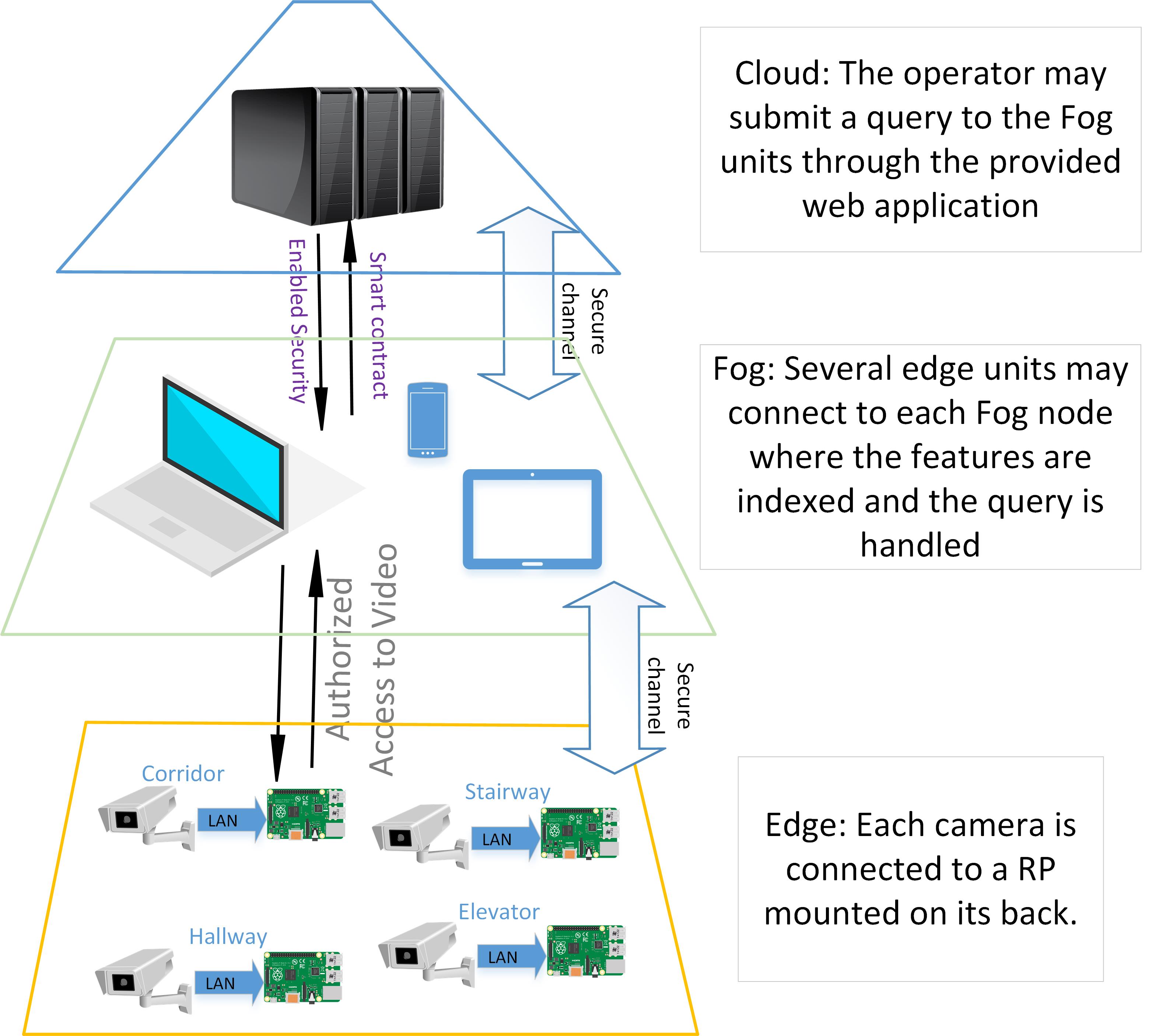}
    \caption{Layered smart surveillance system hierarchy using the edge-fog-cloud computing paradigm.}
    \label{fig:arch}
    \vspace{-10pt}
\end{figure}

%In this section we are going to discuss the basics of our search algorithm, showing the highlights of the search and results presentation protocol in an overview. Then the details of each step is discussed in the subsequent sections. 

\subsection{Hierarchical Platform}

Figure \ref{fig:arch} presents the layered architecture of the proposed I-ViSE system that follows the edge-fog-cloud computing paradigm \cite{nikouei2018real} and \cite{xu2018blendcac}. At the edge, smart cameras are deployed to collect video streams and conduct pre-processing for object detection and feature extraction. Due to its limited computing capability, more complex tasks are deployed on the fog layer. Each fog node communicates and manages several edge devices. Normally, fog nodes are allocated that are close to the geolocation of the associated edge nodes. Meanwhile, fog nodes communicate with the cloud node, accepting dispatched queries.

Due to the attractive features of low cost, small energy consumption, and reasonable computing power; the edge nodes of the I-ViSE system are smart cameras built with the Single Board Computers (SBC), such as Raspberry Pi Model 3 or Model 4. With a good tradeoff between the computing power and energy utility, the edge nodes accommodate microservices that execute video pre-processing and feature extracting tasks. Meanwhile, the fog nodes are expected to be capable of maintaining the throughput required as a middle node. The Fog node may be a tablet or a laptop that is deployed close to the locations of the smart cameras. For instance, the laptop carried on the patrolling vehicle driven by a security officer. The cloud center has connection to all of the edge and fog nodes in the network and can access any device when needed. Human operators can issue queries to all the fog nodes from the cloud center.

%\textcolor{red}{
In this work, the microservices architecture is realized through docker image implementation, which is selected because of many advantages. The docker system is easy to use and it's availability through the cloud connection supports convenient interaction, efficient fetching, and pre-built image processing. Two docker container images are built for the I-ViSE platform, one for the edge nodes and the other for the fog nodes, each running a webservice through the Python’s Flask web-framework.%}

%\textcolor{red}{
Security is derived from protection from attack over hardware, software, and data. While current study assumed robustness from security, future work with leverage our work in (1) software security: authentication and access control, (2) hardware security: temper evident platforms based on the blockchain ledger, and (3) data security: from our work on the context-driven situation awareness in which context features are checked to determine the pragmatic results for consistency.%}

\subsection{Working Flow}
%In the search algorithm, depicted 
As illustrated in Fig \ref{fig:arch}, the edge hierarchy is adopted to connect a huge number of cameras into a tree-based graph to fog nodes, which are in charge of the request handling. The model has several advantages such as good scalability and easy updates and management when needed. This flexible platform architecture can easily handle more cameras when more edge and fog nodes are added in future deployments.

\begin{figure}[t]
    \centering
        \includegraphics[width=0.42\textwidth]{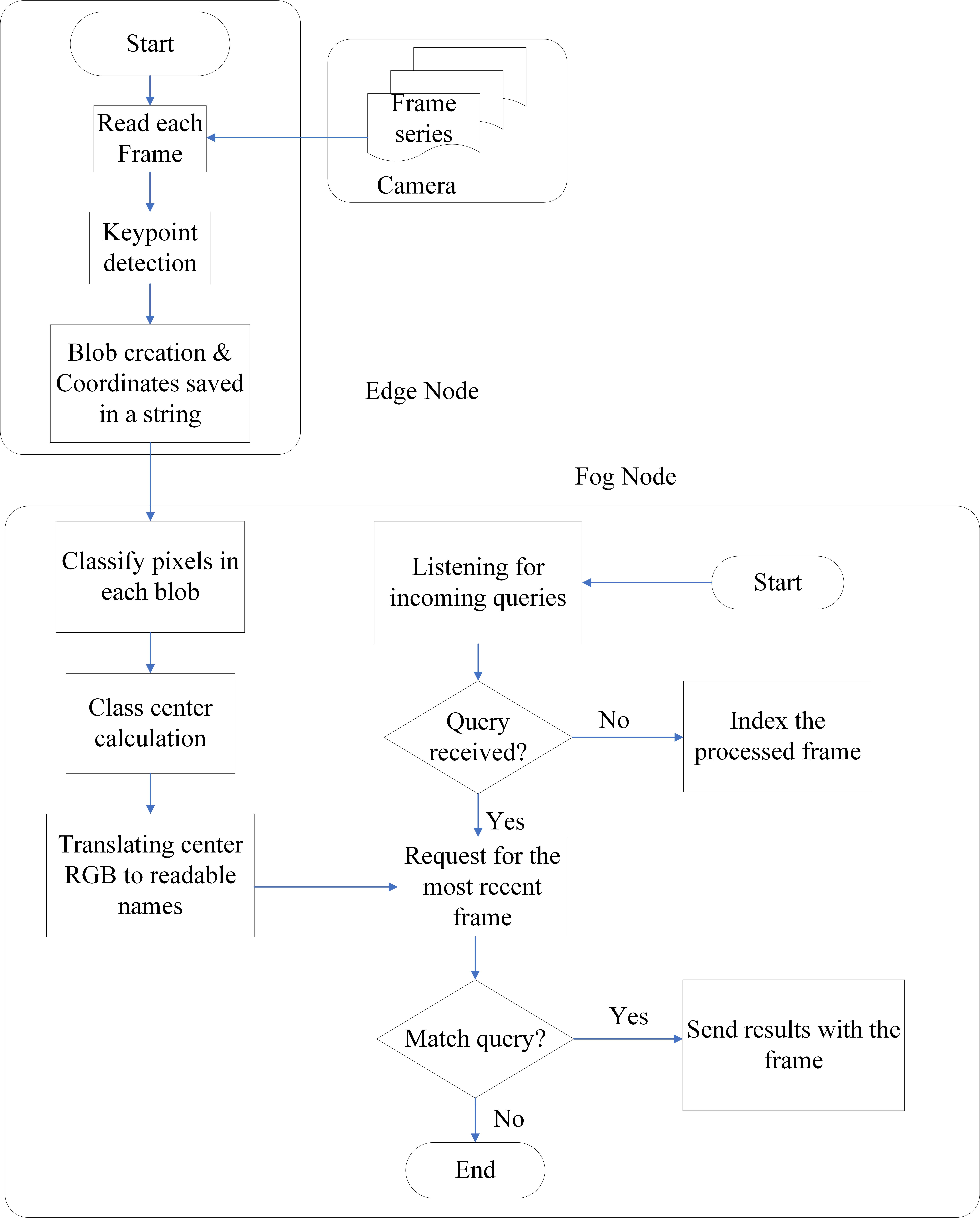}
    \caption{Data Flow Flowchart of the proposed decentralized video query.}
    \label{fig:algorithm}
    \vspace{-10pt}
\end{figure}

Figure \ref{fig:algorithm} shows the workflow of proposed I-ViSE scheme. Once a video frame is captured by the camera, it is streamed to the SBC on-site in the edge node. The SBC accepts every frame from the camera and marks it for either omitting or processing. On receiving a query from the fog or cloud layer, the edge device starts processing the current video frames. According to the keywords provided in the query, the edge node will detect whether or not there is an object of interest in the frames. If the object is detected, the keypoints of the human body and the corresponding colors in the key regions are extracted and sent to the fog node. The query-matching is conducted at the fog node as the edge device cannot handle the computing-intensive task. If there is a match, the fog node reports the results to the operator, including the frame time, the camera ID and the location.

%For a particular job of the real-time search in the video, unless there is a request from the corresponding fog node, there is no need to process the frame, and it can be omitted. For indexing purpose each frame can be processed and the results may be saved in a SQL table in a fog node for the challenges of indexing. Considering the velocity of a human and the fact that this algorithm is primarily considers humans as objects of interest, However, there is no need for the edge node to process each frame. This issue is also being handled indirectly by the better and more computationally powerful hardware releases each year. Moreover, the proposed decentralized approach creates a pipeline where while the previous frame is processed at the fog, a new frame can be handled in the edge. For more details on the performance of the edge please refer to Section \ref{sec:experimental}. 
% (Please refer to Section \ref{sec:experimental}). 

%showing the steps taken in the edge device where the video frame is read, being pre-processed, and being resized and fed to the deep model for human pose detection. The resulting string is then sent to the fog node where it is processed for each of the sections and the unsupervised model assigns a color to each section. The results are then ready for comparison to the query string or indexing. 

\section{Frame Preprocessing at the Edge}
\label{sec:frame}

On-site processing at the edge is the most ideal solution. The video frames are processed immediately once they are collected by the camera, minimizing the communication overhead incurred by the raw video transmission through the network. Although the query is initialized from the operator through the cloud and fog nodes, most of the raw footage data is not relevant. Actually, the useful information can be delivered back to the node that initiated the query using a small amount of bytes, which results from the deep model feature extraction and object of interest cropped frame sections.

The fog node handles the query matching and video retrieval. The results are then reported back to the operator along with the ID of the camera with the detected objects. An unsupervised classification model will give the center of the pixel values containing the sections of interest and the center is translated to human readable color names before report generation at the fog. This process is a computing intensive task accomplished by the fog node reducing the communication traffic and removing the dependence on the remote cloud node.

In this work, a real-time human pose estimation model proposed by researchers from IBM \cite{cao2017realtime} is adopted, which is based on the OpenPose in the TensorFlow framework %\textcolor{red}{
(for more information on the accuracy measurements of OpenPose deep model, please see \cite{cao2018openpose})%}
. As illustrated by Fig. \ref{fig:algorithm}, the edge node feeds the video frames to the Open Pose model to get the human key points. The DNN model is trained on COCO data set with more than 64,000 images for 2D pose estimation. This model is available by IBM, deployed in a docker container. The container removes the need for environment preparations and may receive the frame through the open port. By the implementation of the post-processing, the results are in the format of a string for each frame. On the output of the model, there are two branches; one to give the confidence score in the body-joint being predicted and the part affinity fields for parts association. Each branch has multiple stages of convolutional layers providing a feature map. At the end of each stage the results in the feature map produced in confidence branch is summed up with the resulting feature map from the part affinity fields. 

The part affinity fields present a gradient for each pixel on the human body along and close to the line connecting two body points. The ground truth for $L^{*}_{c,k}(p)$, which is a unit vector that points from one body part to the other along a limb, is described as Eq. (\ref{eq:L}):

\begin{equation}
 L^{*}_{c,k}(p) = \begin{cases}\textbf{v},  & \mbox{if \emph{p} on limb \emph{c}, \emph{k}}  \\0 & \mbox{Otherwise } \end{cases}
 \label{eq:L}
\end{equation}

\noindent where \textbf{\textit{v}} is the unit vector as defined by Eq. (\ref{eq:vector}): 

\begin{equation}
 \textbf{v} = \frac{(x_{j2,k} - x_{j1,k})}{||x_{j2,k} - x_{j1,k}||_2}
 \label{eq:vector}
\end{equation}

\noindent where the points $x_{j2,k}$ and $x_{j1,k}$ represent the limb $c$ of the person $k$. Each point $p$ is a pixel that may be along the limb or not represented by $L^{*}_{c,k}(p)$. The threshold showing if the designated point $p$ is placed on a certain limb $c$ is given as:% Eq. (\ref{eq:treshold}):

\begin{equation}
\begin{split}
 0 \leq v.(p-x_{j1,k})\leq l_{c,k} \\
 0 \leq v_{\perp}.(p-x_{j1,k})\leq \delta_{c,k}
\end{split}
\label{eq:treshold}
\end{equation}

\noindent here the limb width is $\delta_{c,k}$ and the limb length is $l_{c,k}$. 

In this button-up approach, post processing is required after the model gives the results so that the points are grouped for each human. This task is done through grouping the points based on connection between them and the direction of the connection link between each pair of keypoints. The model has 75.6 mean-Average Precision on the COCO test data improving the accuracy of the human gesture estimation in comparison with other models. The approach demonstrates moderate, but manageable, resource consumption on a Raspberry PI board. 

\begin{figure*}[t]
    \centering
        \includegraphics[width=0.73\textwidth]{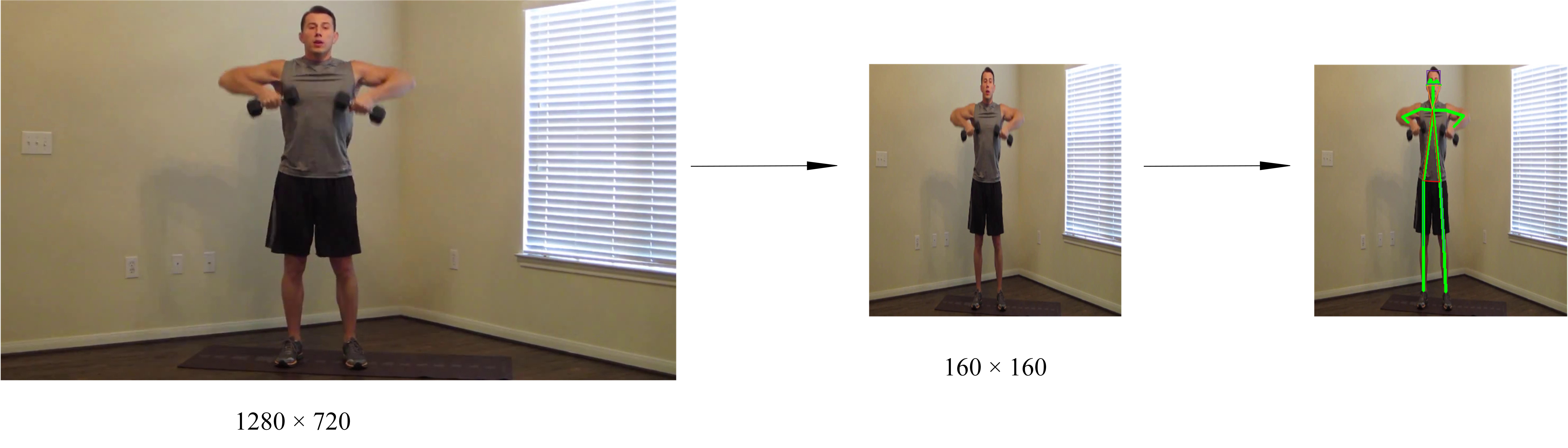}
    \caption{The image resize and keypoint detection. These keypoints are used for part extraction and color detection.}
    \label{fig:bodies}
    \vspace{-10pt}
\end{figure*}

Figure \ref{fig:bodies} shows a sample image that is processed using the DNN model. In Fig. \ref{fig:bodies}, after the frame is captured, the service at the edge node implements frame resizing since $160\times160$ is  the accepted image size which can be fed to this DNN model. Also some other filters smooth the image and reduce noise. These steps improve the accuracy of the DNN model for human keypoint detection. The image in the middle of Fig. \ref{fig:bodies} shows the frame after initial pre-processing. The frame is then processed using another docker container with the DNN implementation. The text results can be shown on the image in form of lines and keypoints as the green lines on the far right section of that figure. The position of the keypoints are of importance to conduct highlighted portions of body for color detection purposes.

One down side to using docker is that the operating system limits the docker containers to prevent system crash, which in return in a smaller device such as our edge node, the execution takes even longer \cite{nikouei2019decentralized}. %\textcolor{red}{
However, the modular capability that the docker containers provide is aligned with the Microservices architecture making scaling easier.%} 

The last step conducted by the edge device is to crop the areas of interest. If $H_{f,c,l}$ shows the point of left hip of the person $c$ in frame sequence $f$, and $H_{f,c,r}$ shows the right hip, connecting them to the lower part of the neck, $N_{f,c}$, a triangle is shaped, which shows the majority of the upper body of the object and can be used for the color of shirt. The next two important keypoints are the ones of knees named $K_{f,c,l}$ and $K_{f,c,r}$. Connecting them to the corresponding left and right hip points results in two lines along the legs of the object in an array of pixels along the path, which can be used for detection of the color of the pants. The Open Pose model similarly gives $E_{f,c,l}$ and $E_{f,c,r}$, which are the left and right ears. Ears connected to the neck point, gives another triangle. This triangle provides the pixels which are mostly in the face area. Considering the human head to fit in a square, the distance between the ears will create that square. Thus the points of interest in each human are $W=(H_{f,c,l},H_{f,c,r},K_{f,c,l},K_{f,c,r},E_{f,c,l}, E_{f,c,r},N_{f,c})$. 

These sections for each human body in the video frame are fed to the query matching algorithm conducted at the fog nodes. Through an unsupervised K-Nearest-Neighbors (KNN) classification algorithm, the color names presented by the pixel values are extracted and the center of the pixels is accurately obtained. Through classifying the pixel density values for each RGB channel, the expected number of the colors are estimated. 

The output from each batch of edge devices are sent to a fog node along with the areas of interest, where the query-matching procedure will be completed and the results will be reported to the operator.

%%%%%%%%%%%%%%%%%%%%%%%%%%%%%%%%%%%%%%%%%%%%%%%%%%%%%%%%%%%%%%%%%%%%%%%%%%%%%%%%%%%%%%%%%%%%%%%%%%%%%%%%%%%%%%%%%%%%%%%%%%%%%%%%%%%%%%%%%%%%%%%%%%%%%%%%%%%%%%%%%%%%%%%%%%%%%%%%%%%%%%%%%%%%%%%%%%%%%%%%%%%%%%%%%%%
\section{Unsupervised Query Matching}
\label{sec:search}

%The connection between our fog node is static rather than dynamic, meaning that edge devices which are connected to a fog are decided by the operator at the system setup and the connection number may not change. The image processing task is recourse depleting and limited number of connections gives the best Quality of Experience (QoE). Moreover, the physical location of the edge and fog may be close for less communication time spent. 

%There are two types of search as mentioned. The real-time search requests for resources after the query is submitted which may take longer times and so index tables are prepared for future searches that may utilize the query languages to look for the results in the tables. We are going to explore both options in this paper. However, the focus is not on the query handling but on the extraction of information from the frame. 

\begin{figure*}[t]
    \centering
        \includegraphics[width=0.82\textwidth]{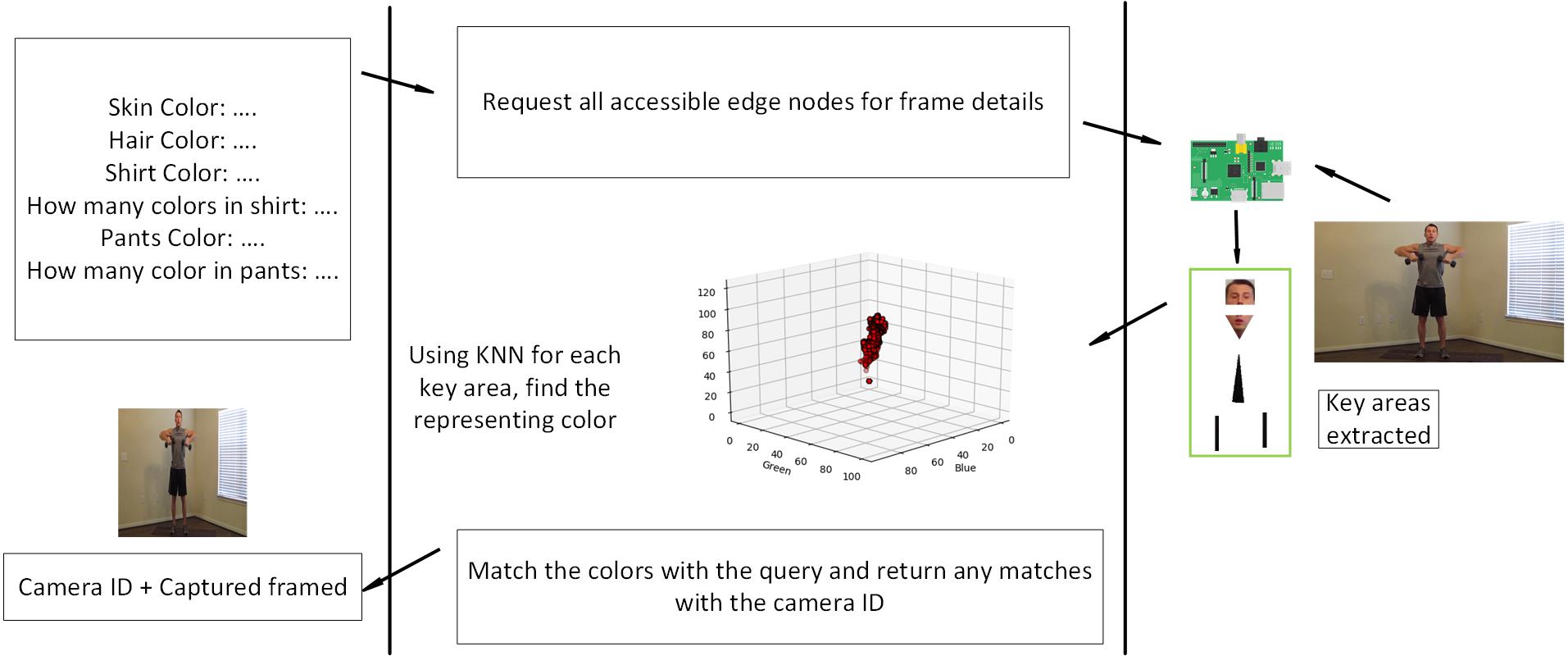}
    \caption{The steps and the working flow of the unsupervised query matching algorithm of I-ViSE.}
    \label{fig:timeline}
    \vspace{-10pt}
\end{figure*}

%With the reception of the query at each edge server, the real-time search for the frame that matches with the query begins at the edge nodes. In this section we discuss the data pipeline. We will also mention indexing features after video processing step for future analytic and historical searches. The focus of this section, is going to be the video processing taking place at the edge node and features to be sent to the corresponding edge server and matching the features with the query. 

Figure \ref{fig:timeline} shows the steps and working flow of the unsupervised query matching algorithm of the I-ViSE scheme. Before the search starts, the algorithm receives a string query with a unique structure from the user. The user submits the query through a cloud node or a fog node, which will communicate with the corresponding edge nodes. The user needs to enter the information they are looking for, such as the number of the colors they are after in each section of the body. For example, the input from the user can be ``blue jeans'', ``red hat'', ``grey T-shirt'', etc. This prevents the user to have access to the public information before having specific description of the person of interest. %blue for the color of the pants 
Grouping pixel values helps with the unsupervised pixel classification, given the number of colors to be expected in each body section. 

The fog node then sends a request to all of the edge nodes that it connects to in order to process the most recent frame that is captured by the camera in an area. On receiving the request from the fog node, the edge nodes feed this frame to its pre-trained DNN, which gives a string showing each of the identifiable people in the frame as well as all of the body joints and their connections. These connections are useful for human pose detection along with the body skeleton. In this paper, these points are leveraged to capture parts of the body and face to allocate the colors the query is interested in. 

Each of the edge nodes sends the body part sections back to the fog node, where all received sections are analyzed. The pixels are translated into a color that can be used to match with the description given by the query. This function is accomplished through a combination of a KNN algorithm and a hash-map data structure.

Each part of the detected human body, as shown in the green rectangle in Fig. \ref{fig:timeline}, is fed to a KNN to identify the color they present. Pixel values are the KNN features. The number of neighborhoods is given by the user for the shirt and pants color implemented with one neighborhood for the face color and one for the hair color. Figure \ref{fig:timeline} shows the pixel values for the grey shirt the person in the sample image is wearing. The kNN clustering representation clearly shows the pixels’ scattering in the body blob of the sample image. Estimating the number of neighborhoods also helps with the noise reduction such as removing a shadow line across the body from which those pixels are considered as outliers and may not change the neighborhood center. 

The center of each neighborhood is the mean of data distribution corresponding to the body section reported in the RGB format. In order for the fog node to compare the results with the query, the last step is to translate the center values to a color name. The colors of the shirt and pants are translated through a 24 hash-map color dictionary where the pixel ranges are mapped to a color names. More detailed names are rarely used in police reports and general colors such as ``red'' or ``blue'' covers a variety of colors. This generalization also reduces the error due to the noise or other light elements that may present a color slightly different. The results are then presented to the operator who can make a final decision. The color map for the face and hair are simple such as ``white'' and ``black'' to present the skin color and ``black'', ``brown'', ``blond'', ``red'' and ``other'' to represent the hair colors.

Finally, the fog node compares the descriptions in the query from the operator to the results of the colors. In case of a match, the frame sequence and the camera ID along with the frame are sent back to the operator. 

\begin{figure}[t]
    \centering
        \includegraphics[width=0.42\textwidth]{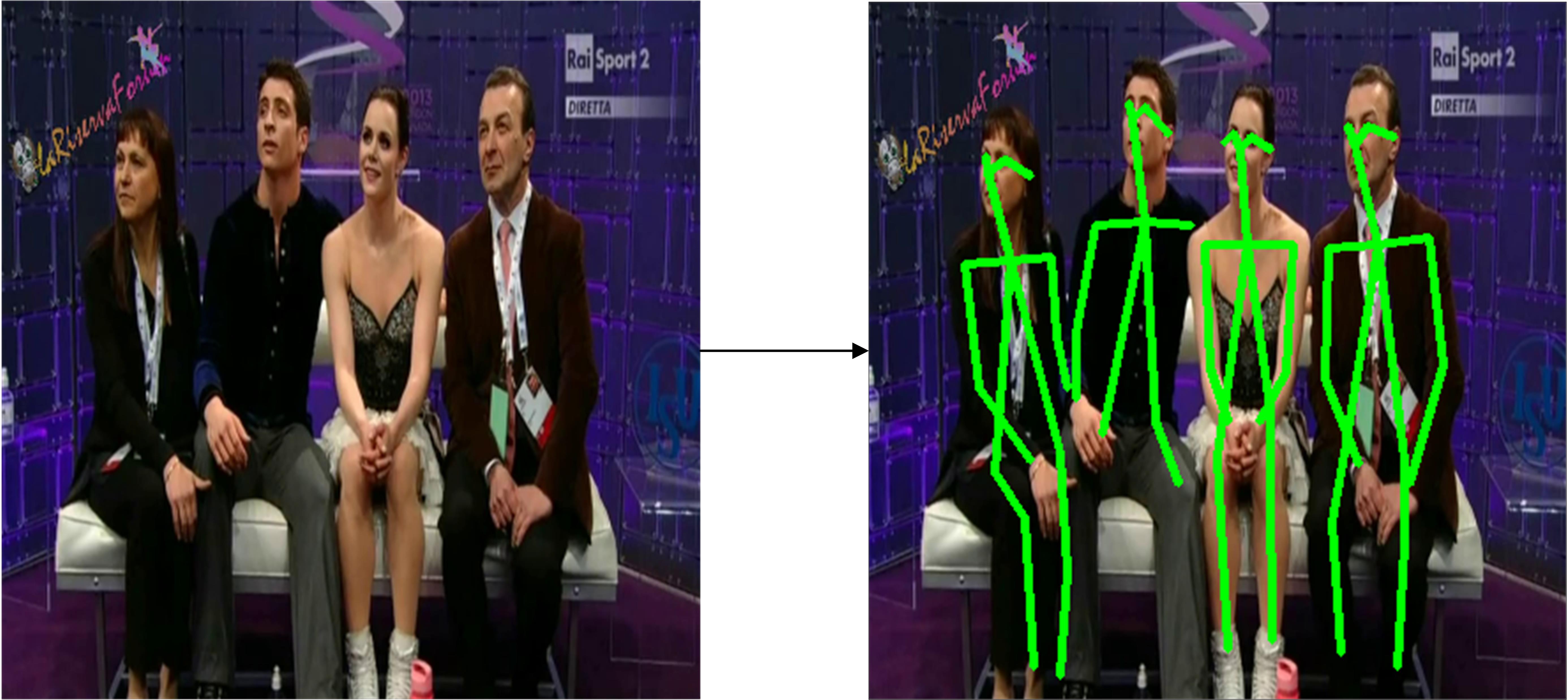}
    \caption{Sample image where the detection CNN misses some parts of some objects of interest.}
    \label{fig:fails}
    \vspace{-10pt}
\end{figure}

The search uncertainty comes from the fact that the DNN model may fail to detect every human and every keypoint in the frame. In case of a missing keypoint, the suspected contour could not be defined and consequently the color of the part could not be retrieved. The model is trained to predict the position of the keypoints. However, this may not be the output if the object of interest (human) has a sharp angle towards the camera. Figure \ref{fig:fails} shows some scenarios where the detection failed. In Fig. \ref{fig:fails} four people are sitting next to each other. The algorithm is confused about the right hand side male's left leg and classified it as the far right lady's right leg. Moreover, the algorithm failed to classify the ears which are not visible from this camera angle. Readers interested in more detailed information are referred to \cite{cao2017realtime}. 

%%%%%%%%%%%%%%%%%%%%%%%%%%%%%%%%%%%%%%%%%%%%%%%%%%%%%%%%%%%%%%%%%%%%%%%%%%%%%%%%%%%%%%%%%%%%%%%%%%%%%%%%%%%%%%%%%%%%%%%%%%%%%%%%%%%%%%%%%%%%%%%%%%%%%%%%%%%%%%%%%%%%%%%%%%%%%%%%%%%%%%%%%%%%%%%%%%%%%%%%%%%%%%%%%

\section{Experimental Results}
\label{sec:experimental}

The accuracy of the I-ViSE scheme is determined by the accuracy of the CNN adopted for object detection. Our experimental study has verified there is not any degradation introduced in the query processing flow. Therefore, the experimental results reported in this section focus on the performance matrix in terms of frame processing speed and utility of computing and communication resources. Table \ref{tabel:per} compares the accuracy of our CNN model with two other state of the art models on the MPII human keypoint detection test.

\subsection{Experimental setup}

As mentioned earlier, Raspberry PI model 4B is adopted as the edge node running Raspbian (Buster) operating system. It is with 4GB LPDDR4-3200 SDRAM and Broadcom BCM2711, Quad core Cortex-A72 (ARM v8) 64-bit SoC @ 1.5GHz chip. The cameras are Logitech 1080p with 60 frames per second connected to the USB port of the Raspberry Pi boards. The fog node is a laptop PC running Ubuntu 16.04 operating system. The PC has a 7th generation Intel core i7 processor @3.1GHz and 32 GB of RAM. The wireless connection between the fog and edge is through wireless local area network (WLAN) with 100 Mbps. The operator can send query through the TCP/IP protocol and is considered to be using the same fog node. Each edge module is handled with a CPU core on the fog (single threaded execution), so that more edge boards can be connected at the same time. Other resource managing software also may be used on top of the platform for better resource management.    

\subsection{Performance Evaluation}

%Video query is a broad topic which is relatively untouched by the smart surveillance community. The concept of video analysis is limited to object detection and classification of pictures, which actually can be considered as the first step toward understanding of the video data. Attempts to query visual data uses the deep learning models to classify frames with the specific objects that are in each frame while giving the bounding box around each one. 

%This platform queries the color of different parts of the humans as objects of interest and returns the results to the operator or who submits the query. Also, the platform is capable of indexing all of the frames that are processed in the fog node connected to the edge. The connections are based on HTML. 

\subsubsection{Preprocessing at the Edge}

\begin{table}

\begin{tabular}{|c|c|c|c|c|c|c|}
\hline
Architecture & Head & Sho & Elb & Wri & Hip & \textbf{mAP} \\
\hline\hline
DeeperCut \cite{insafutdinov2016deepercut} & 78.4 & 72.5 & 60.2 & 51.0 & 57.2 & 59.5\%\\
\hline 
Iqbal et al. \cite{iqbal2016multi} & 58.4 & 53.9 & 44.5 & 35.0 & 42.2 & 43.1\% \\
\hline
Our implementation & 91.2 & 87.6 & 77.7 & 66.8 & 75.4 & 75.6\%\\
\hline
\end{tabular}
\caption{Implemented model for human keypoint extraction accuracy compared to other DL models \cite{cao2017realtime}.}
\label{tabel:per}
\vspace{-25pt}
\end{table}

\begin{figure}[b]
    \centering
    \vspace{-15pt}        \includegraphics[width=0.42\textwidth]{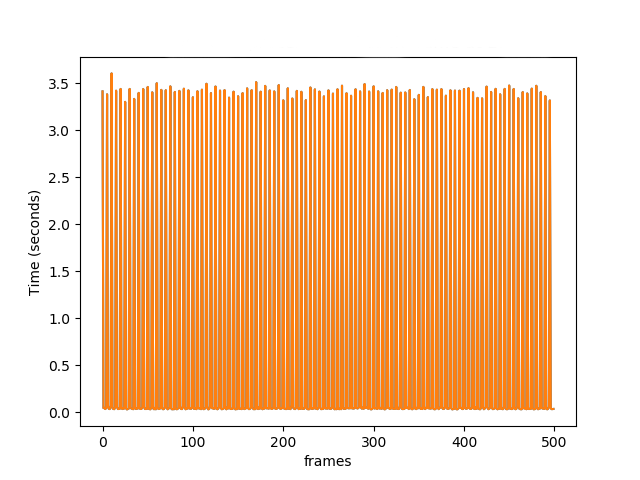}
    \caption{Frame processing delays at the edge.}
    \label{fig:CPU_edge}
\end{figure}

To support real-time, online queries, the most critical link in the information processing chain is the delay incurred at the edge nodes where the frames are processed for key points of the objects. Figure \ref{fig:CPU_edge} shows the delays measured at the edge nodes. Figure \ref{fig:CPU_edge} presents a scenario in which there are four people in the frame. Considering the frame-drop rate of 50\% and the average presence of one objects in each frame, the edge nodes can process 1.4 frames per seconds with frames of 1080p resolution. %The same stepping can be seen here as well were the frames who are not resized and fed to the NN, take much lower time compared to the frames which are processed at the edge. It should be noted that this piece of video has 2-4 people detected for the duration of the video which makes it slower in processing.  

\begin{figure}[t]
    \centering
        \includegraphics[width=0.42\textwidth]{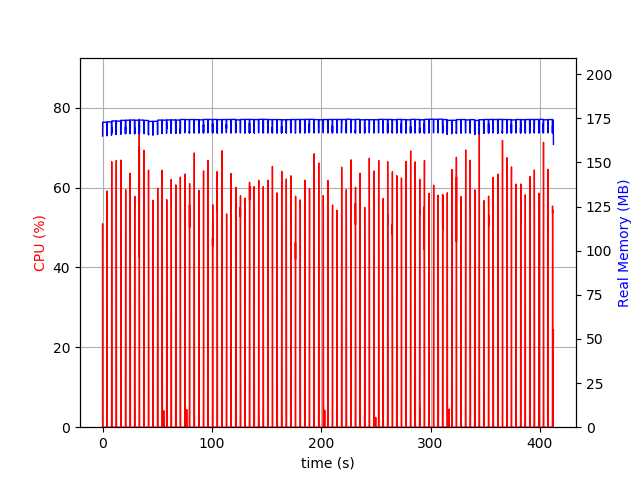}
    \caption{Memory and CPU usage of the edge node (Raspberry PI 4B).}
    \label{fig:speed_edge}
    \vspace{-15pt}
\end{figure}

Figure \ref{fig:speed_edge} presents the CPU and memory utility for processing about 400 seconds long section of a video by an edge node, a Raspberry Pi 4B device,  %As can be seen the algorithm execution is limited to every 5 frames, and the results is the 
which shows a burst usage pattern in the memory and CPU when the CNN is being executed. % usage and then dropping in a pattern. 
The use of 175 MB of the memory and the 80\% CPU gives us the confidence to use a resource constrained Raspberry PI as the edge device to fulfill the needs of the I-ViSE scheme. %The CPU of the device is recorded about ~80\% which is acceptable in long duration of time. 
Meanwhile, 80\% also validates the design that allocates the query matching procedure on the fog side.
%The frame is processed for human pose detection and the results are sent to the fog node. The edge node's capability of video processing proves to be limited as so outsourcing is inevitable. 

%The burden of this process on the edge can be seen in Fig. \ref{fig:CPU_edge}. This figure shows the main reason why the outsourcing needed to take place. Additionally, this figure presents the memory needed for execution of this video processing which is vital in a device that has access to very limited memory.

\subsubsection{Load on the Communication Network}

\begin{figure}[b]
    \vspace{-15pt}
    \centering
        \includegraphics[width=0.45\textwidth]{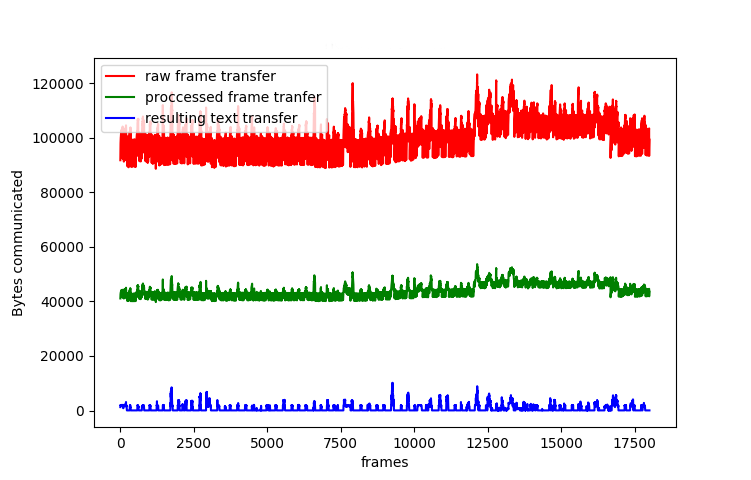}
    \caption{Bytes sent in raw video versus sending the features and frame blobs after processing at the edge.}
    \label{fig:network}

\end{figure}

%It is reminded that the benefit of breaking the algorithm into pieces between different hardware classes as explained in this paper is the prevention of the burden that is video transmission on the network. 
Instead of outsourcing the raw video to the fog node, the I-ViSE edge devices only send the string along with image blobs that can be used by the classifier. If the frame does not include any object of interest, there is no need to transfer any information. This strategy is beneficiary to the communication network. Figure \ref{fig:network} shows the gain in workload. It compares the transmission of the whole frame versus the transmission of the data extracted in a hundred frames period. It shows the instances where there is at least one person visible in the frame, which suggests an even lower average in the overall communication traffic. Figure \ref{fig:network} shows that the traffic is reduced from an average of about 100KB to about 45KB with a $55\%$ reduction in the network traffic resulting from the edge computation. 

\subsubsection{Query Processing at the Fog}

\begin{figure}[t]
    \centering
        \includegraphics[width=0.42\textwidth]{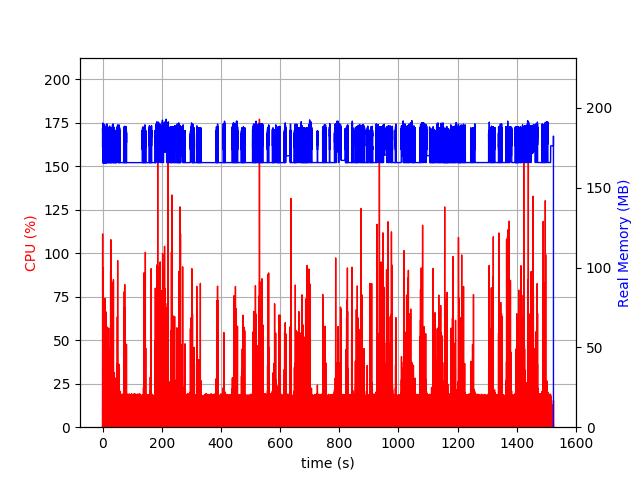}
    \caption{CPU and memory usage of the fog node while processing the feed from one camera on a single thread (Overclocking from the based CPU speed is shown to be higher than 100\%).}
    \label{fig:CPU_fog}
    \vspace{-15pt}
\end{figure}

\begin{figure}[b]
    \vspace{-15pt}
    \centering
        \includegraphics[width=0.4\textwidth]{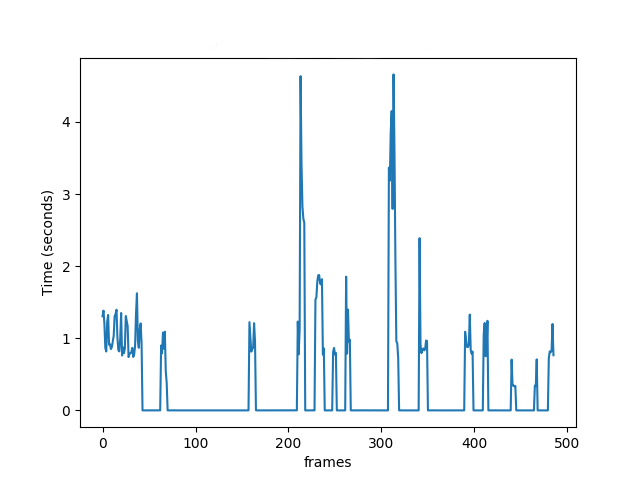}
    \caption{Time required to process each frame at the fog node on single thread.}
    \label{fig:time_fog}

\end{figure}

The experimental results verified that the fog nodes have sufficient capability to handle the query after the results are taken from the edge. Figure \ref{fig:CPU_fog} shows the memory and CPU usage in query processing for a period of almost a thousand seconds. Figure \ref{fig:CPU_fog} is generated using the data available through the \emph{top} application in Unix for system monitoring purposes and recorded through \emph{PSrecord} (same as the edge node). Moreover, the time needed in the fog node to process a single frame for a period of run-time is given in Fig. \ref{fig:time_fog}. Notice that the data shown in Fig. \ref{fig:time_fog} are after the edge has finished the frame preprocessing, so there is no delay related to the fog waiting for the new information to work on. There are several spikes in the processing time corresponding to the number of people in the frame. Such that as the number increases the algorithm requires more time to search and match for each object. The pixel evaluation algorithm takes about 0.9 seconds to classify and translate the pixel values to human readable names and stack it up for a given object on the fog node utilized in this study on a single thread.

\vspace{-20pt}
\subsection{Discussions}

As illustrated by the data flow in Fig. \ref{fig:algorithm}, the query and search processes in the I-ViSE model does not aim at any specific human object in the video frames. Even in a case that an officer searches for a specific target, the model does not reveal any data about the face or clothing type of the object. The data in categorized is based on the color of the shirt, hat, or/and hair. There is not a video frame or image to be outsourced to the data center. Therefore, by design the I-ViSE scheme protects the privacy of the pedestrians who are walking in front of the cameras while provides law enforcement agents the capability to effectively search for a suspect. The I-ViSE scheme does not introduce bias towards any group or ethnicity, as the KNN model is not given a knowledge of the person of interest. The CNN is pre-trained to extract feature artifacts related to the human body and does not provide further information regarding the identification of the subject.

%in no parts of the search process details about the general pedestrian is not shared unless an officer searches for an specific target. Even in that case the model does not share any data about the face or clothing type of the target. The data in categorized based on the color of the clothing and hair. No video or image is transferred to the cloud node which further protects the privacy of the passengers who are walking in front of the camera. The police force, however, gains the ability to effectively search for a suspect. The model introduces no bios towards any group or ethnicity, as the KNN model has no knowledge of the person himself/herself and only is presented with the section of body that is of interest. The CNN is pre-trained to extract feature artifacts related to the human body and provides no further information regarding subject.}

\section{Conclusions}
\label{sec:conclusion}

In this paper, we presented a unique searching algorithm for video querying using a DNN that has the potential of being deployed on the edge architecture. Using the microservices scheme, the proposed I-ViSE platform is divided to simple tasks to reduce communications, improve accuracy, and provide real-time performance. The I-ViSE system is capable of reading real-time video frames and performing the search for a query entry in average of two seconds. I-ViSE also has the capability to create an index table on the fog device for future searches. The platform allows the operator to search through the large scale smart surveillance system video archive with high-level, subjective descriptions, such as the color of clothes or the hair of a human. Through a proof-of-concept prototype utilizing a Raspberry Pi as the edge device, the I-ViSE scheme is validated that achieves the design goals. We hope this effort may inspire more discussions and research in the security surveillance community. 

The I-ViSE is highlighted for man-machine surveillance based on an assumption that the imagery being processed has undergone ``interpretability'' scores to ensure that the images processed contain meaningful content and image quality. The sensor (noise), environment (illumination, weather), and target (movements) influence the performance while the image quality is related to the processing, geometry, and effects. These conditions were held constant in the collections to focus on timeliness. Future studies will show the variations in performance relative to these variations.

%Through a proof-of-concept prototype built on top of Raspberry Pi as the edge device, the I-ViSE scheme is validated that achieves the design goals. Unfortunately, due to the lack of labeled data set or comparable research, we could not provide some comparison studies. We hope this effort may inspire more discussions and research in the security surveillance community.  %who is in the frame and is recognized by the deep model for human pose estimation. 

\ifCLASSOPTIONcaptionsoff
  \newpage
\fi

\bibliographystyle{IEEEtranS}

\bibliography{L-CNN.bib}

% Generated by IEEEtranS.bst, version: 1.14 (2015/08/26)
\begin{thebibliography}{10}
\providecommand{\url}[1]{#1}
\csname url@samestyle\endcsname
\providecommand{\newblock}{\relax}
\providecommand{\bibinfo}[2]{#2}
\providecommand{\BIBentrySTDinterwordspacing}{\spaceskip=0pt\relax}
\providecommand{\BIBentryALTinterwordstretchfactor}{4}
\providecommand{\BIBentryALTinterwordspacing}{\spaceskip=\fontdimen2\font plus
\BIBentryALTinterwordstretchfactor\fontdimen3\font minus
  \fontdimen4\font\relax}
\providecommand{\BIBforeignlanguage}[2]{{%
\expandafter\ifx\csname l@#1\endcsname\relax
\typeout{** WARNING: IEEEtranS.bst: No hyphenation pattern has been}%
\typeout{** loaded for the language `#1'. Using the pattern for}%
\typeout{** the default language instead.}%
\else
\language=\csname l@#1\endcsname
\fi
#2}}
\providecommand{\BIBdecl}{\relax}
\BIBdecl

\bibitem{ananthanarayanan2017real}
G.~Ananthanarayanan, P.~Bahl, P.~Bod{\'\i}k, K.~Chintalapudi, M.~Philipose,
  L.~Ravindranath, and S.~Sinha, ``Real-time video analytics: The killer app
  for edge computing,'' \emph{computer}, vol.~50, no.~10, pp. 58--67, 2017.

\bibitem{cao2018openpose}
Z.~Cao, G.~Hidalgo, T.~Simon, S.-E. Wei, and Y.~Sheikh, ``Openpose: realtime
  multi-person 2d pose estimation using part affinity fields,'' \emph{arXiv
  preprint arXiv:1812.08008}, 2018.

\bibitem{cao2017realtime}
Z.~Cao, T.~Simon, S.-E. Wei, and Y.~Sheikh, ``Realtime multi-person 2d pose
  estimation using part affinity fields,'' in \emph{Proceedings of the IEEE
  Conference on Computer Vision and Pattern Recognition}, 2017, pp. 7291--7299.

\bibitem{carreira2016human}
J.~Carreira, P.~Agrawal, K.~Fragkiadaki, and J.~Malik, ``Human pose estimation
  with iterative error feedback,'' in \emph{Proceedings of the IEEE conference
  on computer vision and pattern recognition}, 2016, pp. 4733--4742.

\bibitem{cavallaro2007privacy}
A.~Cavallaro, ``Privacy in video surveillance [in the spotlight],'' \emph{IEEE
  Signal Processing Magazine}, vol.~2, no.~24, pp. 168--166, 2007.

\bibitem{chen2018cascaded}
Y.~Chen, Z.~Wang, Y.~Peng, Z.~Zhang, G.~Yu, and J.~Sun, ``Cascaded pyramid
  network for multi-person pose estimation,'' in \emph{Proceedings of the IEEE
  Conference on Computer Vision and Pattern Recognition}, 2018, pp. 7103--7112.

\bibitem{dang2019deep}
Q.~Dang, J.~Yin, B.~Wang, and W.~Zheng, ``Deep learning based 2d human pose
  estimation: A survey,'' \emph{Tsinghua Science and Technology}, vol.~24,
  no.~6, pp. 663--676, 2019.

\bibitem{fang2017rmpe}
H.-S. Fang, S.~Xie, Y.-W. Tai, and C.~Lu, ``Rmpe: Regional multi-person pose
  estimation,'' in \emph{Proceedings of the IEEE International Conference on
  Computer Vision}, 2017, pp. 2334--2343.

\bibitem{fitwi2019lightweight}
A.~Fitwi, Y.~Chen, and S.~Zhu, ``A lightweight blockchain-based privacy
  protection for smart surveillance at the edge,'' \emph{arXiv preprint
  arXiv:1909.09845}, 2019.

\bibitem{goodfellow2014generative}
I.~Goodfellow, J.~Pouget-Abadie, M.~Mirza, B.~Xu, D.~Warde-Farley, S.~Ozair,
  A.~Courville, and Y.~Bengio, ``Generative adversarial nets,'' in
  \emph{Advances in neural information processing systems}, 2014, pp.
  2672--2680.

\bibitem{herrera2018smart}
L.~F. Herrera-Quintero, J.~C. Vega-Alfonso, K.~B.~A. Banse, and E.~C. Zambrano,
  ``Smart its sensor for the transportation planning based on iot approaches
  using serverless and microservices architecture,'' \emph{IEEE Intelligent
  Transportation Systems Magazine}, vol.~10, no.~2, 2018.

\bibitem{hidalgo2019single}
G.~Hidalgo, Y.~Raaj, H.~Idrees, D.~Xiang, H.~Joo, T.~Simon, and Y.~Sheikh,
  ``Single-network whole-body pose estimation,'' in \emph{Proceedings of the
  IEEE International Conference on Computer Vision}, 2019, pp. 6982--6991.

\bibitem{insafutdinov2016deepercut}
E.~Insafutdinov, L.~Pishchulin, B.~Andres, M.~Andriluka, and B.~Schiele,
  ``Deepercut: A deeper, stronger, and faster multi-person pose estimation
  model,'' in \emph{European Conference on Computer Vision}.\hskip 1em plus
  0.5em minus 0.4em\relax Springer, 2016, pp. 34--50.

\bibitem{iqbal2016multi}
U.~Iqbal and J.~Gall, ``Multi-person pose estimation with local joint-to-person
  associations,'' in \emph{European Conference on Computer Vision}.\hskip 1em
  plus 0.5em minus 0.4em\relax Springer, 2016, pp. 627--642.

\bibitem{ji2019query}
Z.~Ji, Y.~Ma, Y.~Pang, and X.~Li, ``Query-aware sparse coding for web
  multi-video summarization,'' \emph{Information Sciences}, vol. 478, pp.
  152--166, 2019.

\bibitem{kang2019challenges}
D.~Kang, P.~Bailis, and M.~Zaharia, ``Challenges and opportunities in dnn-based
  video analytics: A demonstration of the blazeit video query engine.'' in
  \emph{CIDR}, 2019.

\bibitem{krylovskiy2015designing}
A.~Krylovskiy, M.~Jahn, and E.~Patti, ``Designing a smart city internet of
  things platform with microservice architecture,'' in \emph{Future Internet of
  Things and Cloud (FiCloud), 2015 3rd International Conference on}.\hskip 1em
  plus 0.5em minus 0.4em\relax IEEE, 2015, pp. 25--30.

\bibitem{nagothu2018microservice}
D.~Nagothu, R.~Xu, S.~Y. Nikouei, and Y.~Chen, ``A microservice-enabled
  architecture for smart surveillance using blockchain technology,'' in
  \emph{2018 IEEE International Smart Cities Conference (ISC2)}.\hskip 1em plus
  0.5em minus 0.4em\relax IEEE, 2018, pp. 1--4.

\bibitem{newell2016stacked}
A.~Newell, K.~Yang, and J.~Deng, ``Stacked hourglass networks for human pose
  estimation,'' in \emph{European conference on computer vision}.\hskip 1em
  plus 0.5em minus 0.4em\relax Springer, 2016, pp. 483--499.

\bibitem{nikouei2019kerman}
S.~Y. Nikouei, Y.~Chen, S.~Song, and T.~R. Faughnan, ``Kerman: A hybrid
  lightweight tracking algorithm to enable smart surveillance as an edge
  service,'' in \emph{2019 16th IEEE Annual Consumer Communications \&
  Networking Conference (CCNC)}.\hskip 1em plus 0.5em minus 0.4em\relax IEEE,
  2019, pp. 1--6.

\bibitem{nikouei2019decentralized}
S.~Y. Nikouei, R.~Xu, Y.~Chen, A.~Aved, and E.~Blasch, ``Decentralized smart
  surveillance through microservices platform,'' in \emph{Sensors and Systems
  for Space Applications XII}, vol. 11017.\hskip 1em plus 0.5em minus
  0.4em\relax International Society for Optics and Photonics, 2019, p. 110170K.

\bibitem{nikouei2018real}
S.~Y. Nikouei, R.~Xu, D.~Nagothu, Y.~Chen, A.~Aved, and E.~Blasch, ``Real-time
  index authentication for event-oriented surveillance video query using
  blockchain,'' in \emph{2018 IEEE International Smart Cities Conference
  (ISC2)}.\hskip 1em plus 0.5em minus 0.4em\relax IEEE, 2018, pp. 1--8.

\bibitem{sun2017compositional}
X.~Sun, J.~Shang, S.~Liang, and Y.~Wei, ``Compositional human pose
  regression,'' in \emph{Proceedings of the IEEE International Conference on
  Computer Vision}, 2017, pp. 2602--2611.

\bibitem{wang2013beyond}
F.~Wang and Y.~Li, ``Beyond physical connections: Tree models in human pose
  estimation,'' in \emph{Proceedings of the IEEE Conference on Computer Vision
  and Pattern Recognition}, 2013, pp. 596--603.

\bibitem{wang2019asymmetric}
H.~Wang, C.~Deng, J.~Yan, and D.~Tao, ``Asymmetric cross-guided attention
  network for actor and action video segmentation from natural language
  query,'' in \emph{Proceedings of the IEEE International Conference on
  Computer Vision}, 2019, pp. 3939--3948.

\bibitem{wei2016convolutional}
S.-E. Wei, V.~Ramakrishna, T.~Kanade, and Y.~Sheikh, ``Convolutional pose
  machines,'' in \emph{Proceedings of the IEEE Conference on Computer Vision
  and Pattern Recognition}, 2016, pp. 4724--4732.

\bibitem{xu2018blendcac}
R.~Xu, Y.~Chen, E.~Blasch, and G.~Chen, ``Blendcac: A blockchain-enabled
  decentralized capability-based access control for iots,'' in \emph{the IEEE
  International Conference on Blockchain, Selected Areas in IoT and
  Blockchain}.\hskip 1em plus 0.5em minus 0.4em\relax IEEE, 2018.

\bibitem{xu2018real}
R.~Xu, S.~Y. Nikouei, Y.~Chen, S.~Song, A.~Polunchenko, C.~Deng, and
  T.~Faughnan, ``Real-time human object tracking for smart surveillance at the
  edge,'' in \emph{the IEEE International Conference on Communications (ICC),
  Selected Areas in Communications Symposium Smart Cities Track}.\hskip 1em
  plus 0.5em minus 0.4em\relax IEEE, 2018.

\bibitem{xu2016discovery}
X.~Xu, T.~M. Hospedales, and S.~Gong, ``Discovery of shared semantic spaces for
  multiscene video query and summarization,'' \emph{IEEE Transactions on
  Circuits and Systems for Video Technology}, vol.~27, no.~6, pp. 1353--1367,
  2016.

\bibitem{yang2011articulated}
Y.~Yang and D.~Ramanan, ``Articulated pose estimation with flexible
  mixtures-of-parts,'' in \emph{CVPR 2011}.\hskip 1em plus 0.5em minus
  0.4em\relax IEEE, 2011, pp. 1385--1392.

\bibitem{yu2018survey}
D.~Yu, Y.~Jin, Y.~Zhang, and X.~Zheng, ``A survey on security issues in
  services communication of microservices-enabled fog applications,''
  \emph{Concurrency and Computation: Practice and Experience}, p. e4436, 2018.

\bibitem{zhu2017multi}
X.~Zhu, Y.~Jiang, and Z.~Luo, ``Multi-person pose estimation for posetrack with
  enhanced part affinity fields,'' in \emph{ICCV PoseTrack Workshop}, vol.~3,
  2017, p.~7.

\end{thebibliography}

%\begin{biography}[{\includegraphics[width=1in,height=1.25in,clip,keepaspectratio]{mshell}}]{Michael Shell}

%\begin{IEEEbiography}%[{\includegraphics[width=1in,height=1.25in,clip,keepaspectratio]{picture}}]{John Doe}
%\blindtext
%\end{IEEEbiography}


%\vfill


\end{document}